\pdfoutput=1

\documentclass[11pt]{article}

\usepackage{acl}
\usepackage{times}
\usepackage{latexsym}

\usepackage[T1]{fontenc}

\usepackage[utf8]{inputenc}

\usepackage{microtype}

\usepackage{stfloats}
\usepackage{amsmath}
\usepackage{amsbsy}
\usepackage{amssymb}
\usepackage{graphicx}
\usepackage[export]{adjustbox}
\usepackage{subcaption}
\usepackage{footnote}
\usepackage{multirow}
\makesavenoteenv{tabular}
\makesavenoteenv{table}
\usepackage{url}
\usepackage{tikz}
\usetikzlibrary{fit,positioning}
\usepackage{booktabs}
\usepackage{tabularx}
\usepackage{makecell}
\usepackage[ruled,vlined]{algorithm2e}

\DeclareMathOperator*{\KL}{KL}
\DeclareMathOperator*{\Attention}{Attention}
\DeclareMathOperator*{\MHA}{MHA}
\DeclareMathOperator*{\TransEnc}{Enc}
\DeclareMathOperator*{\TransDec}{Dec}
\DeclareMathOperator*{\FF}{F}
\DeclareMathOperator*{\SA}{SA}
\DeclareMathOperator*{\CA}{CA}

\DeclareMathOperator*{\ARTransDec}{\overline{Dec}}
\DeclareMathOperator*{\QKVDec}{QKVDec}
\DeclareMathOperator*{\ARQKVDec}{\overline{QKVDec}}
\DeclareMathOperator*{\Categorical}{Categorical}
\DeclareMathOperator*{\Concat}{Cat}
\DeclareMathOperator*{\SoftPlus}{SoftPlus}
\DeclareMathOperator*{\softmax}{Softmax}

\DeclareMathOperator*{\ELBo}{ELBo}
\expandafter\def\expandafter\UrlBreaks\expandafter{\UrlBreaks
  \do\a\do\b\do\c\do\d\do\e\do\f\do\g\do\h\do\i\do\j%
  \do\k\do\l\do\m\do\n\do\o\do\p\do\q\do\r\do\s\do\t%
  \do\u\do\v\do\w\do\x\do\y\do\z\do\A\do\B\do\C\do\D%
  \do\E\do\F\do\G\do\H\do\I\do\J\do\K\do\L\do\M\do\N%
  \do\O\do\P\do\Q\do\R\do\S\do\T\do\U\do\V\do\W\do\X%
  \do\Y\do\Z}

\usepackage{soulutf8}
\usepackage[normalem]{ulem}
\usepackage{enumitem}
\setlist{nolistsep}
\usepackage{microtype}

\expandafter\def\expandafter\UrlBreaks\expandafter{\UrlBreaks
  \do\a\do\b\do\c\do\d\do\e\do\f\do\g\do\h\do\i\do\j%
  \do\k\do\l\do\m\do\n\do\o\do\p\do\q\do\r\do\s\do\t%
  \do\u\do\v\do\w\do\x\do\y\do\z\do\A\do\B\do\C\do\D%
  \do\E\do\F\do\G\do\H\do\I\do\J\do\K\do\L\do\M\do\N%
  \do\O\do\P\do\Q\do\R\do\S\do\T\do\U\do\V\do\W\do\X%
  \do\Y\do\Z}

\usepackage{soulutf8}
\usepackage[normalem]{ulem}
\usepackage{xcolor}

\usepackage{enumitem}
\setlist{nolistsep}
\usepackage{microtype}

\iftrue
    \usepackage[final]{proofing}
  \renewcommand\hl[1]{{#1}}  
   {\draftnote{\red{#2}}}
   \newcommand\redHL[1]{}
  \newcommand\todo[1]{}
  \newcommand\barre[1]{}

  \newcommand{\Djame}[1]{}
  \newcommand{\ghazi}[1]{}

\newcommand{\gfcmt}[1]{}
\newcommand{\gfcorr}[2]{#2}
\newcommand{\jlcmt}[1]{}

\newcommand{\dscmt}[1]{}

\else  

\usepackage[draft]{proofing}

\newcommand{\gfcmt}[1]{\textcolor{orange}{#1}}
\newcommand{\gfcorr}[2]{\textcolor{orange}{#1 $\longrightarrow$ #2}}
\newcommand{\jlcmt}[1]{\textcolor{red}{#1}}

\newcommand{\dscmt}[1]{\textcolor{brown}{#1}}

\newcommand{\ghazi}[1]{\textbf{\textcolor{orange}{\hl{Ale: #1}}}}

\newcommand\red[1]{{\color{red}{\bf #1}}}
\newcommand{\barre}[1]{
{\textcolor{red}{{\sout{#1}}}}}

 \newcommand\redHL[1]{\red{\hl{#1}}}
\let\olddraftnote\draftnote
\renewcommand\draftnote[1]{\olddraftnote{\red{#1}}}

\fi

\usepackage[english]{babel} 

\title{Exploiting Inductive Bias in Transformers for Unsupervised Disentanglement of Syntax and Semantics with VAEs}

\author{Ghazi Felhi,\hspace{5px} Joseph Le Roux\\
LIPN\\ Université Sorbonne Paris Nord - CNRS UMR 7030 \\
F-93430, Villetaneuse, France \\
\texttt{\{felhi, leroux\}@lipn.fr} \\\And 
Djamé Seddah \\
INRIA Paris \\ Paris, France \\
\hspace{5px} \texttt{djame.seddah@inria.fr} \\}

\begin{document}
\maketitle
\begin{abstract}

We propose a generative model for text generation, which exhibits disentangled latent representations of syntax 
and semantics.
  Contrary to previous work, this model does not need syntactic information such as constituency parses, or semantic 
  information such as paraphrase pairs.
  Our model relies solely on the inductive bias found in attention-based architectures such as Transformers.

  In the attention of Transformers, \emph{keys} handle information selection while \emph{values} specify what information is conveyed.
  Our model, dubbed QKVAE, uses Attention in its decoder to read latent variables where one latent variable infers keys 
  while another infers values.

  We run experiments on latent representations and experiments on syntax/semantics transfer which show that QKVAE displays clear signs of disentangled syntax and semantics.
  We also show that our model displays competitive syntax transfer capabilities when compared to supervised models and that comparable supervised models need a fairly large amount of data (more than 50K samples) to outperform it on both syntactic and semantic transfer.
  The code for our experiments is publicly available\footnote{\gfcorr{A URL pointing to a public repository will be included in 
  the final version of our submission.}{\href{https://github.com/ghazi-f/QKVAE}{github.com/ghazi-f/QKVAE}}}.

\end{abstract}

\section{Introduction}

Disentanglement, a process aimed at obtaining neural representations with identified meaning, is a crucial component of research
 on interpretability \cite{Rudin2022InterpretableChallenges}. A form of disentanglement that received a lot of interest from the
  NLP community is the separation between syntax and semantics in neural representations \cite{Chen2019ARepresentations, Bao2020, 
  Zhang2020Syntax-infusedGeneration, Chen2020ControllableExemplar, Huang2021GeneratingPairs, Huang2021DisentanglingModelsb}.
Previous works perform disentanglement using paraphrase pairs as information for semantics, and/or constituency parses as information
 for syntax.
The dependence of models on labeled data is known to entail high cost (see \citealp{Seddah2020BuildingHell} on syntactic 
annotation), and to often require new labels to handle problems such as concept drift  \cite{Lu2019LearningReview} and domain adaptation 
 \cite{Farahani2021AAdaptation}.
 
In light of the above, we propose an unsupervised model which directs syntax and semantics into different 
neural representations without semantic or syntactic information. In the Transformer architecture~\cite{Vaswani2017}, the attention 
mechanism is built upon a \emph{query} from a set $Q$, which pools \emph{values} $V$ through \emph{keys} $K$. For each query, 
values are selected according to their matching score computed by the similarity between their corresponding keys and 
the query. Building on an analogy between the $(K, V)$ couple and syntactic roles with their lexical realizations (explicited in 
\S \ref{Behavior}) we present QKVAE\footnote{A contraction of the $(Q, K, V)$ triplet with the VAE acronym.}, a Transformer-based 
Variational Autoencoder (VAE; \citealp{Kingma2014Auto-encodingBayes}).

To build our model, we modify a previous Transformer-based VAE, called the Attention-Driven VAE (ADVAE;
 \citealp{Felhi2021TowardsRoles}).
Using Cross-Attention, our model encodes sentences into two latent variables: $z^{sem}$ to infer values for $V$, and $z^{syn}$ to
 assign keys in $K$ for values in $V$.
These keys and values are then used in the Attention mechanism of a Transformer Decoder to generate sentences.
We show that $z^{syn}$ tends to contain syntactic information, while $z^{sem}$ tends to represent semantic information.
Additionally, comparisons with a supervised model show that it needs a considerable amount of data to outperform our model
 on syntactic and semantic transfer metrics.

Our contributions can be summarized as follows:
\begin{itemize}
    \item We describe QKVAE, a model designed to disentangle syntactic information from semantic information by using separate 
    latent variables for keys and values in Transformers Attention.
    \item We run experiments on a dataset for English which empirically show that the two types of latent variables have strong preferences respectively for syntax and semantic.
    \item We also show that our model is capable of transferring syntactic and semantic information between sentences by using their 
    respective latent variables.
    Moreover, we show that our model's syntax transfer capabilities are competitive with supervised models
    when they use their full training set (more than 400k sentences), and that a supervised model needs a fairly large amount 
    of labeled data (more than 50k samples) to outperform it on both semantic and syntactic transfer.
\end{itemize}

\section{Related Work}
We broadly divide works on explainability in NLP into two research directions.
The first seeks \emph{post hoc} explanations for black-box models, and led to a rich literature of observations on the behavior
 of Neural Models in NLP~\cite{tenney-etal-2019-bert, jawahar-etal-2019-bert, Hu2020AModels,Kodner2020OverestimationModels, Marvin2020TargetedModels,
Kulmizev2020DoFormalisms, rogers-etal-2020-primer}.
Along with these observations, this line of works also led to numerous advances in methodology concerning, for instance,
the use of attention as an explanation~\cite{Jain2019AttentionExplanation,Wiegreffe2020AttentionExplanation}, the validity
of probing~\cite{Pimentel2020Information-TheoreticStructure}, or contrastive evaluation with minimal pairs~\cite{Vamvas2021OnEvaluation}.
The second research direction on explainability in NLP seeks to build models that are explainable by design.
This led to models with explicit linguistically informed mechanisms such as the induction of grammars (RNNG;
 \citealp{Dyer2016RecurrentGrammars}, URNNG; \citealp{Kim2019UnsupervisedGrammars}) or constituency trees (ON-LSTM; 
 \citealp{Shen2019OrderedNetworks}, ONLSTM-SYD; \citealp{Du2020ExploitingApproach}).

Disentangled representation learning is a sub-field of this second research direction which aims at separating neural
 representations into neurons with known associated meanings.
This separation was performed on various characteristics in text such as style \cite{John2020DisentangledTransfer,
 Cheng2020ImprovingGuidance}, sentiment and topic \cite{Xu2020OnSupervision}, or word morphology \cite{Behjati2021InducingAttention}.
In works on disentanglement, consequent efforts have been put in the separation between syntax and semantics, whether merely to obtain 
an interpretable specialization in the embedding space \cite{Chen2019ARepresentations, Bao2020, ravfogel-etal-2020-unsupervised, Huang2021DisentanglingModelsb}, or for
 controllable generation \cite{Zhang2020Syntax-infusedGeneration, Chen2020ControllableExemplar, Huang2021GeneratingPairs, hosking-lapata-2021-factorising, li-etal-2021-deep-decomposable, hosking-etal-2022-hierarchical}.
However, all these works rely on syntactic information (constituency parses and PoS tags) or semantic information (paraphrase pairs).
To the best of our knowledge, our work is the first to present a method that directs syntactic and semantic information into assigned
embeddings in the challenging unsupervised setup.

From a broader machine learning perspective, using knowledge of the underlying phenomena in our data, we design our model QKVAE 
with an inductive bias that induces understandable behavior in an unsupervised fashion.
Among the existing line of applications of this principle \cite{Rezende2016UnsupervisedImages, Hudson2018, 
Locatello2020Object-centricAttention, Tjandra2021UnsupervisedRepresentation}, ADVAE  \cite{Felhi2021TowardsRoles}, the model on
 which QKVAE is based, is designed to separate information from the realizations of different syntactic roles without 
 supervision \gfcorr{$\emptyset$}{on a dataset of regularly structured sentences.}

\section{Background}
In this section, we go over the components of our model, namely VAEs, attention in Transformers, and ADVAE, the model on which QKVAE is based.
\subsection{VAEs as Language Models}
Given a set of observations $w$, VAEs are a class of deep learning models that train a generative model $p_\theta(w) = \int_z p(z) p_\theta(w|z)dz$, where $p(z)$ is a prior distribution on latent variables $z$ that serve as a seed for generation, and $p_\theta(w|z)$ is called the decoder and generates an observation $w$ from each latent variable value $z$.
Since directly maximizing the likelihood $p_\theta(w)$ to train a generative model is intractable, an approximate inference distribution $q_\phi(z|w)$, called the encoder, is used to formulate a lower-bound to the exact log-likelihood of the model, called the Evidence Lower-Bound (ELBo):
\begin{multline}
    \log p_\theta(w) \geq\\ \mathbb{E}_{(z) \sim q_\phi(z|w)}\left[ \log p_\theta(w|z) \right] -\\
    \KL[q_\phi(z|w)||p(z)] = \ELBo(w;z) \label{ELBOEQ}
\end{multline} 
Early works on VAEs as language models have shown that, contrary to non-generative sequence-to-sequence \citep{Sutskever2014b} models, they learn a smooth latent space \citep{Bowman2016GeneratingSpace}. 
In fact, this smoothness enables decoding an interpolation of latent codes (\textit{i.e.} a homotopy) coming from two sentences
 to yield a well-formed third sentence that clearly shares characteristics (syntactic, semantic\dots) with both source sentences.
This interpolation will be used as a control baseline in our experiments.

\subsection{Attention in Transformers.}
\label{ATTBG}
The inductive bias responsible for the disentanglement capabilities of our model is based on the design of Attention in Transformers \cite{Vaswani2017}.
In attention mechanisms, each element of a series of query vectors $Q=\{q_1, \dots, q_{|Q|}\}$
performs a soft selection of values $V=\{v_1, \dots, v_{|V|}\}$ whose  compatibility with the query is given by their corresponding key vector in $K=\{k_1, \dots, k_{|V|}\}$ via dot product.
For each $q_i \in Q$, the series of dot products is normalized and used as weights for a convex interpolation of the values.
Formally, the result is compactly written as:
\begin{align}
\label{SELFATTEQ}
    \Attention(Q, K, V) = \softmax(QK^T)V
\end{align}

Here, we stress that $K$ is only capable of controlling what information is 
selected from $V$, while $V$ is responsible for the value of this information.
Using the above operators and the embedding level concatenation operator $\Concat$,
  Multi-Head Attention ($\MHA$) in Transformers is defined as follows:
 \begin{align*}
     &\MHA(\tilde{Q}, \tilde{K}, \tilde{V}) = \Concat(head_1, ...head_H)W^O \hskip 2mm \\
    &\hskip 2mm s.t :\hskip 1mm head_i = \Attention(\tilde{Q}W_i^Q, \tilde{K}W_i^K, \tilde{V}W_i^V)
 \end{align*}
 Where $W^O$, $W_i^Q$, $W_i^K$, and $W_i^V$ are trainable parameter matrices.
 In turn, Self-Attention ($\SA$) and Cross-Attention
  ($\CA$) are defined, for sets of elements called source $S$ and target
   $T$, as follows:
 \begin{align*}
  \SA(T) = \MHA(T, T, T)\\
  \CA(T, S) = \MHA(T, S, S)
 \end{align*}

  The above $\SA$ mechanism is used to exchange information between elements of target $T$, while in $\CA$, targets $T$ pull
   (or \textit{query} for) information from each element of the source $S$.
  Transformer Encoders ($\TransEnc$) are defined as the composition of layers each consisting of an attention followed by a 
  Feed-Forward Network $\FF$:\footnote{We omit residual connections and layer normalizations after each $\SA$ or $\CA$ for 
  simplicity.}

  \[
    \TransEnc(T) =\tilde{T}_{D^{enc}}, \ \text{s.t. } \tilde{T}_{d} =
    \left \{
      \begin{array}{l}
        T \text{ if } d=0,\text{ else:}\\
        \FF(\SA(\tilde{T}_{d-1}))
      \end{array}
      \right.
  \]

  Transformer Decoders ($\TransDec$) are defined with instances of $\SA$, $\CA$ and $\FF$:
  \[
    \TransDec(T, S) =\tilde{T}_{D^{dec}}, \text{s.t. }:\ 
    \]
  \[ \tilde{T}_{d}  = 
    \left \{
      \begin{array}{l}
       T \text{ if } d=0,\text{ else:}\\
       \FF(\CA(\SA(\tilde{T}_{d-1}),S))
      \end{array}
      \right.
  \]

where $D^{enc}$ and $D^{dec}$ above are respectively the number of layers of $\TransEnc$ and $\TransDec$.
For autoregressive decoding, \citet{Vaswani2017} define a version of $\TransDec$ we will call $\ARTransDec$.
In this version, the result of each $QK^T$ (Eq. \ref{SELFATTEQ}) in Self-Attention is masked so that each $t_i$ in $T$ only queries 
for information from $t_j$ with $j\leq i$.
Even though $\ARTransDec$ yields a sequence of length equal to that of target $T$, in the following sections we will consider 
its output to be only the last element of $\tilde{T}_{D^{dec}}$ in order to express auto-regressive generation 
in a clear manner.

\subsection{ADVAE}
ADVAE is a Variational Autoencoder for unsupervised disentanglement of sentence representations.
It mainly differs from previous LSTM-based \citep{Bowman2016GeneratingSpace} and Transformer-based
 \citep{Li2020Optimus:Space} VAEs in that it uses Cross-Attention to encode and decode latent variables, which is the cornerstone of our model.
In ADVAE, Cross-Attention is used to: \textit{i)} encode information from sentences into a fixed number of
vectorial latent variables; \textit{ii)} decode these vectorial latent variables by using them as sources for the target  sentences generated by a Transformer Decoder.

Formally, let us define $M^\mu$, $M^\sigma$, and $M^w$ to be linear layers that will respectively be used to obtain the latent variables' means and standard deviations, and the generated words' probabilities, $L$  the number of vectorial latent variables $z=\{z_1, \dots, z_{L}\}$,
and finally $E=\{e_1,\dots, e_{L}\}$ and $D=\{d_1,\dots, d_{L}\}$ two sets of $L$ trainable embeddings.
Embeddings $e_{i}$ and $d_{i}$ serve as fixed identifiers for the latent variable $z_{i}$ respectively in the encoder and in the decoder.

Given input token sequence $w$, the encoder $q_\phi(z|w)=\prod_{l} q_{\phi}(z_{l}|w)$ first yields parameters $\mu_l$ and $\sigma_l$ to be used by the diagonal Gaussian distribution of each of the latent variables $z_l$ as follows\footnote{To simplify equations, we omit word embedding look-up tables and positional embeddings.}:

\begin{flalign}
    \Tilde{z} = \TransDec(e; \TransEnc(&w))&& \nonumber\\
     \forall\hskip 1mm l \hskip 2mm\text{ s.t. }\hskip 1mm 1\leq l\leq& L: &&\nonumber\\
    \mu_{l}=  M^\mu(&\Tilde{z}_l), \hskip 2mm \sigma_{l} =  \SoftPlus(M^\sigma(\Tilde{z}_l)) &&\nonumber\\
     z_l \sim \mathcal{N}(\mu&_l; \sigma_l)&& \label{ADVAEncEq}
\end{flalign}

Cross-Attention is also used by the ADVAE decoder to dispatch information from the \textit{source} latent variable samples to the 
\textit{target} generated sequence. Accordingly, using a beginning-of-sentence token $w_0$,
  $p_\theta(w|z)=\prod_{i}p_\theta(w_{i}|w_{<i},z)$ yields probabilities
for the categorical distribution of the generated tokens $w$ by decoding latent variables $z$ concatenated with their embeddings $d$:
\begin{flalign*}
  y = \Concat&(d; z)\nonumber\\
\forall\hskip 1mm i \hskip 2mm\text{ s.t. }\hskip 0mm &1\leq i\leq |w|: \hskip 2mm\\
    &\Tilde{w}_i =\ARTransDec(w_0, \dots, w_{i-1};
    \TransEnc(y)) \nonumber\\
    &w_i \sim \Categorical(\softmax(M^w(\Tilde{w}_i))) \nonumber
\end{flalign*}



\section{QKVAE: Using separate latent variables for Keys and Values}
In this section, we describe the architecture of our model, the behavior it entails, and  how we deal with the optimization 
challenges it poses.

\begin{table*}[t]
    \small
    \centering
    \begin{tabularx}{15.5cm}{X X p{1cm} X X c p{2.1cm} p{5cm}}
    \cline{1-5} 
    \cline{1-5} 
      $v$ & child & to wear & cloak & winter & & & \\
      
      $k1$ & nsubj & root & dobj & $\emptyset$ &$\longrightarrow$ & decoded ($v$, $k1$): & A child wears a cloak.\\
    
      $k2$ & agent & root &  nsubjpass & pobj  &$\longrightarrow$ & decoded ($v$, $k2$):& A cloak is worn, in winter, by a child\\
    \cline{1-5}     \cline{1-5} 
      \end{tabularx}
      \caption{Example of interpretable values for the $v$ and $k$ in our model with $L=4$.
We display a sentence transiting from the active form to the passive form, to illustrate how different \textit{keys} arranging the 
same \textit{values} can lead to the same minimal semantic units being rearranged according to a different syntactic structure.
We also stress that a different set of \textit{keys} may omit or bring forth an element from the \textit{values} vector 
(\textit{e.g.} "winter" here above).}
    \label{tab:QKexpect}
\end{table*} 
\subsection{QKVAE architecture}

The modification we bring to ADVAE is aimed at controlling how information is selected from the latent space with the value of a
 newly introduced latent variable.
We call this latent variable $z^{syn}$, and refer to the latent
 variables already formulated in ADVAE as  $z^{sem}=\{z^{sem}_1, \dots, z^{sem}_{L}\}$.
$z^{syn}$ is obtained with the same process as each $z^{sem}_l$ (Eq.
\ref{ADVAEncEq}), \textit{i.e.} by adding an additional
 identifier embedding $e_{s}$, and matrices $M^{\mu s}$ and $M^{\sigma s}$ to obtain its mean and standard-deviation parameters.
 
For the QKVAE Decoder, we modify the Transformer Decoder $\TransDec$ into $\QKVDec$ so as to use Multi-Head Attention with separate
inputs for keys and values instead of Cross-Attention :

 \[
  \hskip -10mm \QKVDec(T; S_K; S_V) =\tilde{T}_{D^{QKV}}, \ \text{s.t. :}
\] 
\[\tilde{T}_{d} =
  \left \{
    \begin{array}{l}
      T \text{ if } d=0,\text{ else:}\\
      \FF(\MHA(\SA(\tilde{T}_{d-1}), S_K, S_V)
    \end{array}
    \right.
\]
where $D^{QKV}$ is the number of layers.
Similar to $\ARTransDec$, we define $\ARQKVDec$ to be the auto-regressive version of $\QKVDec$.
The QKVAE decoder yields probabilities for the generated tokens by using this operator on values given by $z^{sem}$ concatenated with
embeddings $d$, and keys given by a linear transformation on $z^{syn}$: 
\begin{flalign}
  v = \Concat(d;& z^{sem})\nonumber, \hskip 2 mm k = M^s(z^{syn})\nonumber\\
\forall\hskip 1mm i\hskip 2mm\text{ s.t. }\hskip 2mm 1&\leq i\leq |w|: \hskip 2mm\nonumber\\
    \Tilde{w}_i =& \ARQKVDec(w_0, \dots, w_{i-1};k ; v) \nonumber\\
w_i \sim& \Categorical(\softmax(M^w(\Tilde{w}_i))) \nonumber
\end{flalign}

where $M^s$ is a linear layer.\footnote{The output of $M^s$ is reshaped to obtain a matrix of keys.}
While ADVAE already uses Cross-Attention to encode and decode latent variables,
our model uses separate variables to obtain keys and values for Multi-Head Attention in its decoder.

\subsection{QKVAE Behavior}
\label{Behavior}
In the \gfcorr{Cross-Attention}{Multi-Head Attention} of our decoder, $z^{syn}$ controls keys, and $z^{sem}$ controls values.
In other words, the value of each 
$z^{sem}_l$ is called to be passed to the target sequence according to its key which is given by the variable $z^{syn}$.
Therefore, given a query, $z^{syn}$ decides which content vector $z^{sem}_l$ participates most to the value of the generated token 
at each generation step.
To better get a gist of the kind of behavior \emph{intended} by this construction, we assume in Table \ref{tab:QKexpect} for 
explanatory purposes, that our decoder has one layer and one attention head, that the value of each $k^l$  in key matrices $k_1$
and $k_2$ corresponds to syntactic roles, and that each $v^l$ informs on the realization of the corresponding syntactic role.
 Table \ref{tab:QKexpect} displays the resulting sentence when each of $k1$ and $k2$ are coupled with $v$.

In the examples in Table \ref{tab:QKexpect}, the generator uses a query at each generation step to pick a word in a manner 
that would comply with English syntax. Therefore, the key of each value should inform on its role in the target structure, which 
justifies syntactic roles as an adequate meaning for keys.
 
Although our model may stray from this possibility and formulate non-interpretable values and keys, keys will still inform on the \emph{roles} of values in the target structure, 
and therefore \gfcorr{serve as a schedule for the injection of the values}{influence the way values are injected } 
into the target sequence.
And given the fact that our model uses multiple layers and attention
heads and the continuous nature of keys in Attention (as opposed to discrete syntactic role labels), our model performs a multi-step and
continuous version of the behavior described in Table \ref{tab:QKexpect}.

Injecting values into the structure of a sentence requires the decoder to model this structure.
Previous works have shown that
this is well within the capabilities of Transformers.
Specifically, \citet{Hewitt2019ARepresentations} showed that 
Transformers embed syntactic trees in their inner representations, \citet{Clark2019WhatAttentionb}
showed that numerous attention heads attend to specific syntactic roles, and \gfcorr{\citet{Felhi2021TowardsRoles}}{we \cite{Felhi2021TowardsRoles}}
showed that Transformer-based VAEs can capture the realizations of syntactic roles in latent variables obtained with Cross-Attention.

\subsection{Balancing the Learning of $z^{sem}$ and $z^{syn}$}
Similar to ADVAE, we \gfcorr{$\emptyset$}{ use a standard Normal distribution as a prior $p(z)=p(z^{sem})p(z^{syn})$ and} 
 train QKVAE with the $\beta$-VAE objective
 \cite{Higgins2019-VAE:Framework} which is simply $\ELBo$ (Eq.
\ref{ELBOEQ}) with a weight $\beta$ on its Kullback-Leibler ($\KL$) term.
\citet{Higgins2019-VAE:Framework} show that a higher $\beta$ leads to better unsupervised disentanglement.
However, the $\KL$ term is responsible for a phenomenon called \emph{posterior collapse} where the latent variables
 become uninformative
 and are not used by the decoder \cite{Bowman2016GeneratingSpace}.
Therefore, higher values for $\beta$ cause poorer reconstruction performance \cite{Chen2018c}.
To avoid posterior collapse, we follow \citet{Li2020AText}: \textit{i)} We pretrain our model as an autoencoder
by setting $\beta$ to 0; \textit{ii)} We linearly increase $\beta$ to its final value ($\KL$ annealing; 
\citealp{Bowman2016GeneratingSpace}) and we threshold each dimension of the $\KL$ term with a factor $\lambda$ (Free-Bits
strategy; \citealp{Kingma2016ImprovedFlow}).

In preliminary experiments with our model, we observed that it tends to encode sentences using only $z^{sem}$.
As \gfcorr{$z^{syn}$ and $z^{sem}$ are independent}{ we use conditionally independent posteriors\footnote{These posteriors are ADVAE encoders (Eq. \ref{ADVAEncEq}).} 
$q(z^{syn}|w)$ and $q(z^{sem}|w)$ for our latent variables}, their $\KL$ terms (Eq.
\ref{ELBOEQ}) can be written seperately, and they can therefore be weighted separately with different values of $\beta$.
Using a lower $\beta$ for $z^{syn}$ as was done by \cite{Chen2020ControllableExemplar}
\gfcorr{$\emptyset$}{\footnote{Although not explicitly mentioned in the paper, this is performed in their companion source code.}}
  did not prove effective in making it informative for the model.
Alternatively, linearly annealing $\beta$ for $z^{sem}$ before $z^{syn}$ did solve the issue. This \gfcorr{$\emptyset$}{intervention on the learning process} 
was inspired by the work of \citet{Li2020ProgressiveRepresentations} which shows that latent variables used at different parts
 of a generative model should be learned at different paces.

\section{Experiments}
\subsection{Setup}
\paragraph{Data}
To compare our model to its supervised counterparts, we train it with data from the English machine-generated paraphrase pairs dataset 
ParaNMT \cite{Wieting2018ParanMT-50M:Translations}.
More specifically, we use the 493K samples used by 
\citet{Chen2020ControllableExemplar}\footnote{\href{https://drive.google.com/open?id=1HHDlUT\_-WpedL6zNYpcN94cLwed\_yyrP}
  {https://drive.google.com/open?id=1HHDlUT\_-WpedL6zNYpcN94cLwed\_yyrP}} to train their model VGVAE.
Since our model is unsupervised, we only use the reference sentences (half the training set) to train our model. Using the development and test sets of
 ParaNMT, \citet{Chen2020ControllableExemplar} also provide a curated set of triplets formed by a target sentence (\emph{target}),
   a semantic source (\emph{sem\_src}),and a syntactic source (\emph{syn\_src}).
The semantic source is a paraphrase of the target sentence, while the syntactic source is selected by finding a sentence that is syntactically close to the target (\textit{i.e.} edit distance between the sequence of PoS Tags of both sentences is low\footnote{We follow \citet{Chen2020ControllableExemplar} by using this evaluation data, although edit distance between PoS tags might not be a good proxy for syntactic similarity.}) and semantically different from the paraphrase (has low BLEU score with it).
Contrary to paraphrases in the training set of ParaNMT, paraphrases from this set were manually curated.
These triplets are divided into a development set of 500 samples and a test set of 800 samples.
We display results on the test set in the main body of the paper. The results on the development set, which lead to the same conclusions, are reported in Appendix \ref{DEVRES}.

\paragraph{Training details \& hyper-parameters}
Encoders and Decoders in QKVAE are initialized with parameters from BART \cite{Lewis2020BART:Comprehension}. After
manual trial and error on the development set, we set 
the sizes of $z^{syn}$ and $z^{sem}$ to 768, and $L$ to 4. Further Hyper-parameters are in Appendix \ref{HPAPPEN}.
\gfcorr{$\emptyset$}{We train 5 instances of our model and report the average scores throughout all experiments.}
\paragraph{Baselines}
We compare our system to 4 previously published models, where 2 are supervised and 2 are unsupervised: 
\textit{i) VGVAE \cite{Chen2020ControllableExemplar}: } a VAE-based paraphrase 
generation model with an LSTM 
architecture. This model is trained using paraphrase pairs and PoS Tags to separate syntax and semantics into two latent variables.
This separation is used to separately specify semantics and syntax to the decoder in order to produce paraphrases; \textit{ii) SynPG 
\cite{Huang2021GeneratingPairs}:} A paraphrase generation Seq2Seq model based on a Transformer architecture
 which also separately encodes syntax and semantics for the same purpose as VGVAE.
This model is, however, trained using only source sentences with their syntactic parses, without paraphrases; \textit{iii) Optimus 
\cite{Li2020Optimus:Space}:} A large-scale VAE based on a fusion between BERT \cite{Devlin2018b} and GPT-2
 \cite{Radford2018LanguageLearners} with competitive performance on various NLP benchmarks; \textit{iv) ADVAE: } This model is QKVAE
 without its syntactic variable. \gfcorr{$\emptyset$}{The size of its latent variable is set to 1536 to equal the total size of latent variables in QKVAE.}

Official open-source instances\footnote{
VGVAE: \href{https://github.com/mingdachen/syntactic-template-generation/}{github.com/mingdachen/syntactic-template-generation/};
SynPG: \href{https://github.com/uclanlp/synpg}{github.com/uclanlp/synpg};
Optimus: \href{https://github.com/ChunyuanLI/Optimus}{github.com/ChunyuanLI/Optimus};
ADVAE: \href{https://github.com/ghazi-f/ADVAE}{github.com/ghazi-f/ADVAE}
}
 of the 4 models above are available,
 which ensures accurate comparisons. The off-the-shelf instances of 
VGVAE and SynPG are trained on ParaNMT with GloVe\footnote{Gains could be observed with better embeddings for supervised models, 
but we stick to the original implementations.}\cite{Pennington2014} embeddings.
We fine-tune a pre-trained Optimus on our training set following instructions from the authors. Similar to our model,
 we initialize ADVAE with 
parameters from BART\cite{Lewis2020BART:Comprehension} and train  \gfcorr{$\emptyset$}{5 instances of} it on ParaNMT with $L=4$.
\subsection{Syntax and Semantics Separation in the Embedding Space}
We first test whether $z^{syn}$ and $z^{sem}$ respectively specialize in syntax and semantics.
A syntactic (resp. semantic) embedding should place syntactically (resp. semantically) similar sentences close to each other in the embedding space.

Using the (\textit{target, sem\_src, syn\_src}) triplets, we calculate for each embedding the probability that \textit{target} is closer
 to \textit{sem\_src} than it is to \textit{syn\_src} in the embedding space.
For simplicity, we refer to the syntactic and semantic embeddings of all models as $z^{syn}$ and $z^{sem}$.
For Gaussian latent variables, we use the mean parameter as a representation (respectively the mean direction parameter from 
the von Mises-Fisher distribution of the semantic variable of VGVAE).
We use an L2 distance for Gaussian variables and a cosine distance for the others. Since Optimus and ADVAE do not have separate 
embeddings for syntax and semantics \textit{i)} We take the whole embedding for Optimus; \textit{ii)}\gfcorr{We choose the highest (resp. lowest)
 scoring latent variable $z_l$ (Eq. \ref{ADVAEncEq}) for ADVAE as a semantic (resp. syntactic) variable}{For ADVAE, we 
 measure the above probability on the development set for each latent variable $z_l$ (Eq. \ref{ADVAEncEq}).
 Then, we choose the latent variable that places \emph{target} sentences closest 
 to their \emph{sem\_src} (resp. \emph{syn\_src}) as a semantic (resp. syntactic) variable}.
The results are presented in Table \ref{enc_res}.

\begin{table}\small
\centering
\begin{tabular}{l c c}
\hline
  &$z^{sem} \uparrow$ & $z^{syn} \downarrow$\\
\hline\multicolumn{3}{c}{\textit{Supervised Models}}\\\hline
VGVAE & 99.9& 14.8\\
SynPG & 93.4& 26.5\\
\hline\multicolumn{3}{c}{\textit{Unsupervised Models}}\\\hline
Optimus & 91.8 & -\\
ADVAE & 39.5 & 40.0\\
QKVAE & 89.2& 26.4\\

\hline
\end{tabular}
\caption{\label{enc_res} \gfcmt{I replaced the results of ADVAE and QKVAE with the 5 runs averages here. }
The probability*100 that an embedding places a target sentence closer to its semantic source than it is to its syntactic 
source in the embedding space. Arrows ($\uparrow$/$\downarrow$) indicate whether higher or lower scores are better.}
\end{table}

\begin{table*}[t]
  \small
\centering
\begin{tabular}{l c c c|| c c c|| c c c }
\hline
&\multicolumn{3}{c}{\textit{sem\_src}} & \multicolumn{3}{c}{\textit{syn\_src}}& \multicolumn{3}{c}{\textit{target}}\\
&\emph{STED}$\uparrow$ & \emph{TMA2}$\downarrow$ & \emph{TMA3}$\downarrow$ & 
\emph{STED}$\downarrow$ & \emph{TMA2}$\uparrow$ & \emph{TMA3}$\uparrow$ &
\emph{STED}$\downarrow$ & \emph{TMA2}$\uparrow$ & \emph{TMA3}$\uparrow$ \\
\hline\multicolumn{10}{c}{\textit{Control and Reference baselines}}\\ 
\hline
\textit{sem\_src} &0.0 & 100 & 100 & 13.0& 40.3& 4.8 & 12.0& 39.6& 7.0 \\
\textit{syn\_src} & 13.0& 40.3& 4.8& 0.0& 100& 100& 5.9& 84.3& 45.8\\
Optimus & 11.6& 50.0& 15.9& 9.2& 61.6& 23.6& 10.2& 58.9& 21.8\\
\hline\multicolumn{10}{c}{\textit{Supervised Models}}\\
\hline
VGVAE & 13.1& 39.9& 5.4& 3.3& 86.4& 64.1& 6.7& 80.4& 44.6\\
SynPG & 11.7& 41.9& 18.0& 13.5& 74.1& 10.5& 13.1& 69.1& 13.3\\
\hline\multicolumn{10}{c}{\textit{Unsupervised Models}}\\\hline
ADVAE    & 11.9& 47.3& 14.0& 10.3& 54.3$^\dag$& 19.2$^\dag$& 11.1& 52.3& 17.0\\
QKVAE    & 12.7& 40.2& 7.8& 7.2& 68.2& 39.5& 8.9& 63.9& 28.1\\
\hline
\end{tabular}
\caption{\gfcmt{I replaced the results of ADVAE and QKVAE with the 5 runs averages here. }Syntactic transfer results. \emph{STED} is the Syntactic Tree Edit Distance, and \emph{TMA2/3} is the exact matching
 between constituency trees truncated at the $2^{nd}$/$3^{rd}$ level.\label{syn_res}}
\end{table*}

Table \ref{enc_res} clearly shows for QKVAE, SynPG, and VGVAE that the syntactic (resp. semantic) variables lean towards positioning 
sentences in the embedding space according to their syntax (resp. semantics).
Surprisingly, the syntactic variable of our model specializes \gfcorr{considerably more in syntax 
(\textit{i.e.} has lower score) than}{in syntax (\textit{i.e.} has low score) as much as} that of SynPG.
The generalist latent variable of Optimus seems to position sentences in the
latent space according to their semantics. Accordingly, we place its score in the $z^{sem}$ column. Interestingly, the variables in
 ADVAE have very close scores and \gfcorr{show}{score well below 50, which shows} that the entire ADVAE embedding leans more towards syntax. This means
that, without the key/value distinction in the Attention-based decoder, the variables specialize more in
 structure than in content.

\subsection {Syntactic and Semantic Transfer}
Similar to \citep{Chen2020ControllableExemplar}, we aim to produce sentences that take semantic content from \emph{sem\_src} 
sentences and syntax from \emph{syn\_src} sentences.
For each of SynPG, VGVAE, and QKVAE we simply use the syntactic embedding of \emph{syn\_src}, and the semantic embedding of 
\emph{sem\_src} as inputs to the decoder to produce new sentences. Using the results of the specialization test in the previous experiment,
we do the same for ADVAE by taking the 2 latent variables that lean most to semantics (resp. syntax) as semantic (resp. syntactic) variables.
The output sentences are then scored in terms of syntactic and semantic similarity with \emph{sem\_src}, \emph{syn\_src} and \emph{target}.

\begin{table}[t]
  \small
\centering
\begin{tabular}{p{0.045\textwidth} c| c|| c| c|| c| c}
\hline
&\multicolumn{2}{c}{\textit{sem\_src}} & \multicolumn{2}{c}{\textit{syn\_src}}& \multicolumn{2}{c}{\textit{target}}\\
& \emph{M}$\uparrow$&\emph{PB}$\uparrow$
& \emph{M}$\downarrow$&\emph{PB}$\downarrow$
& \emph{M}$\uparrow$& \emph{PB}$\uparrow$ \\
\hline
\multicolumn{7}{c}{\textit{Control and Reference baselines}}
\\
\hline
\textit{sem\_src}    & 100 & 1.0 &  6.9& 0.14& 28.8&  0.84\\
\textit{syn\_src}    & 6.9& 0.14&  100 & 1.0 &  12.1& 0.16\\
Optimus & 12.4& 0.34& 15.9& 0.39&  10.8 & 0.32\\
\hline
\multicolumn{7}{c}{\textit{Supervised Models}}\\\hline
VGVAE    &17.6 & 0.58&  15.3& 0.18& 24.9 & 0.58\\
SynPG  & 45.9 & 0.87& 8.0& 0.13& 25.2& 0.75\\
\hline
\multicolumn{7}{c}{\textit{Unsupervised Models}}\\
\hline
ADVAE & 8.0& 0.19& 8.3$^\dag$& 0.17& 7.4& 0.19\\
QKVAE & 12.8& 0.35&  11.0& 0.19 & 12.6 &0.34\\
\hline
\end{tabular}
\caption{\gfcmt{I replaced the results of ADVAE and QKVAE with the 5 runs averages here. }Semantic transfer results. \emph{M} is the Meteor score, and \emph{PB} is the ParaBart cosine similarity.\label{sem_res}}
\end{table}
 
\paragraph{Control and reference baselines} Beside model outputs, \gfcorr{we produce scores for \emph{syn\_src} and \emph{sem\_src}}{
  we also use our syntactic and semantic comparison metrics, explicited below, to compare \emph{syn\_src} and \emph{sem\_src} sentences to one
   another and to  \emph{target} sentences}. 
Additionally, using Optimus, we embed \emph{sem\_src} and \emph{syn\_src}, take the dimension-wise average of both embeddings, 
and decode it.
As VAEs are known to produce quality sentence interpolations \cite{Bowman2016GeneratingSpace, Li2020Optimus:Space}, 
the scores for this sentence help contrast a naïve fusion of features in the embedding space with a composition of well 
identified disentangled features.

\paragraph{Transfer metrics} We measure the syntactic and semantic transfer from source sentences to output sentences.
\textit{i) Semantics:} For semantics, previous works \cite{Chen2020ControllableExemplar, Huang2021GeneratingPairs} rely on 
lexical overlap measures such as BLEU \cite{Papineni2001BLEU:Translation}, ROUGE \cite{lin2004rouge}, and Meteor
 \cite{denkowski-lavie-2014-meteor}. As will be shown in our results, the lexical overlap signal does not capture semantic 
transfer between sentences when this transfer is too weak to produce paraphrases. Therefore, we use Meteor (\emph{M}) in conjunction with ParaBART \cite{Huang2021DisentanglingModelsb} a model where BART \cite{Lewis2020BART:Comprehension} is fine-tuned using syntactic information to produce neural representations that represent maximally semantics and minimally syntax.
We measure the cosine similarity between sentences according to ParaBART embeddings (\emph{PB}). \textit{ii) Syntax:} We use the script of \cite{Chen2020ControllableExemplar} to produce a syntactic tree edit distance (STED) between the constituency trees of sentences, as was done to assess VGVAE.
Additionally, following the evaluation procedure designed by \citet{Huang2021GeneratingPairs} for SynPG, we measure the Template
 Matching Accuracy 
between sentences, where the template is the constituency tree truncated at the second level (TMA2).
TMA2 is the percentage of sentence pairs where such templates match exactly.
We extend this measure by also providing it at the third level (TMA3).
Results are presented in Tables \ref{syn_res} and \ref{sem_res}. \gfcorr{$\emptyset$}{In both Tables, the comparison 
scores between sentences and \emph{syn\_src} that are not significantly\footnote{We consider differences to be significant 
if their associated $t$-test yields a $p$-value<0.01.} different from the same scores produced with regard to 
\emph{sem\_src} are marked with $^\dag$.}

\paragraph{Sanity checks with metrics and baselines} We notice in Table \ref{sem_res} that using Meteor as a semantic similarity measure results in various inconsistencies.
For instance, paraphrases \textit{target} have a higher Meteor score with the syntactic
sources than with interpolations from \textit{Optimus}.
It can also be seen that the Meteor score between outputs from VGVAE and both syntactic and semantic sources are rather close
\gfcorr{$\emptyset$}{\footnote{This was not observed by \citet{Chen2020ControllableExemplar}, as they only compared outputs 
from VGVAE to the target paraphrases.}}.
In contrast, ParaBART score behaves as expected across comparisons in Table~\ref{sem_res}.
Consequently, we retain ParaBART score as a semantic similarity measure. 
In the following, we use the scores between \textit{sem\_src}, \textit{syn\_src}, and
\textit{target} (first two rows in Tables \ref{sem_res} and \ref{syn_res}) as reference scores for unrelated sentences, paraphrase pairs, and syntactically similar sentences.  
\paragraph{Comparing the supervised baselines} 
VGVAE and SynPG greatly differ in scores.
It can be seen that SynPG copies a lot of lexical items from its semantic input \gfcorr{$\emptyset$}{(high Meteor score)} which allows for higher semantic similarity scores.
However, Table \ref{syn_res} shows that SynPG transfers syntax from \textit{syn\_src} at a high level (high TMA2, but low TMA3). 
In contrast, VGVAE transfers syntax and semantics in a balanced way and achieves the best syntax transfer scores overall 
\gfcorr{$\emptyset$}{(lowest STED with \emph{syn\_src} and \emph{target})}.

\paragraph{Analysing the scores of QKVAE} The semantic similarity scores \textit{PB} of QKVAE outputs with  \textit{target} and \textit{sem\_src} are close to those of Optimus outputs.
Although these scores are low compared to supervised models, they are notably higher
 than semantic similarity scores between unrelated sentences (\textit{e.g.} \textit{syn\_src} and \textit{sem\_src}). 
However, in contrast to Optimus, QKVAE outputs \gfcorr{are rather semantically unrelated to \textit{syn\_src}}{display low PB scores 
with \textit{syn\_src}, which show that they draw very little semantic information from the syntactic sources}. 
Concerning syntactic transfer in Table \ref{syn_res}, QKVAE outputs
share syntactic information with \textit{syn\_src} on all levels (low STED, and high TMA2 and TMA3).
Our model is even \gfcorr{on par}{competitive} with SynPG on TMA2, and better on TMA3 and STED.
As expected, the scores comparing QKVAE outputs to \textit{sem\_src} show that they \gfcorr{are syntactically unrelated sentences}{
  share very little syntactic information}.
On the other hand, ADVAE shows poor transfer performance on syntax and semantics, with only slight differences
between scores w.r.t \textit{syn\_src} and scores w.r.t \textit{sem\_src}.

\begin{table*}[t]
  \small
  \centering
  \begin{tabularx}{16cm}{|X|X|X|X|X|X|}
  \hline
   \emph{sem\_src}& \emph{syn\_src}& SynPG & VGVAE& QKVAE& \emph{target}\\
  \hline \hline
we have destroyed the 49th armored division.&
	concomitant usage is not recommended.&
  	we have destroyed the 49th armored division.&
    	armored division hasn't destroyed.&
      	this military force will be destroyed. &
        	49th armored division has been destroyed .\\\hline																
let the fire burn and put a piece of hot iron in it.&
	sing a song. sing a song for boys.&
  	don't put the fire in it burn a hot piece of iron and fire.&
    	burn the fire. put the iron on burns.&
      	come on fire. get a fire on it.&
        	keep this fire going. keep a piece of hot iron on it.\\\hline																				
they took the lunch boxes ?&
	have you given me your hands ?&
  	do they boxes took the lunch ?&
    	have they taken them your snacks ?&
      	have you heard of some lunch ?&
        	have they taken the lunch boxes ?	\\\hline																		
does it have a coach ?&
	that's a phone switcher, right ?&
  	how does it have a coach ?&
    	that's a coach coach, right ?& 
      	that's a warden, huh?&
        	it has a coach, no ?\\\hline																			
an old lady in a cemetery.&
	that is a bad time for a war.&
  	there's a lady in an old cemetery.&
    	that's an old lady in the cemetery.&
      	this is a strange place for a woman.&
        	there is an old lady in the cemetery.\\\hline
        	
don't be afraid.&
    there are still many places to go.&
        you don't be afraid.&
            there aren't be afraid to be.&
                there will be no need to worry.&
                    there is no need to be afraid .\\\hline
                    
isn't there a door open ?&
	the machines are still good, right ?&
  	a isn't open door there ?&
    	the doors aren't open, right ?&
      	the door will be open, okay?&
        	there is a door open, right ?\\\hline
   \end{tabularx}
  \caption{\gfcmt{I replaced the previous examples with ones from the new runs, and put 7 instead of 3.} Syntactic sources (\emph{syn\_src}), semantic sources (\emph{sem\_src}), the sentences produced when using them with different models, and the corresponding correct paraphrases (\emph{target}).}
  \label{tab:qualiRes}
\end{table*} 
\subsection{Comparing our Model to a Supervised Model with Less Data}
Since VGVAE displays balanced syntactic and semantic transfer capabilities,
 we use it for this experiment where we train it on subsets of sizes in $\{10K, 25K, 50K, 100K\}$ 
from its original training data.
Our goal is to find out how much labeled data is needed for VGVAE to outperform our
unsupervised model on both transfer metrics.
 
\begin{figure}[!h]
  \hskip -0.6cm
  \begin{minipage}[b]{\textwidth}
  \begin{adjustbox}{minipage=\textwidth,scale=0.59}
  \hspace{ 1cm} \includegraphics[trim={0.7cm 0.2cm 1.5cm 1.2cm},clip] {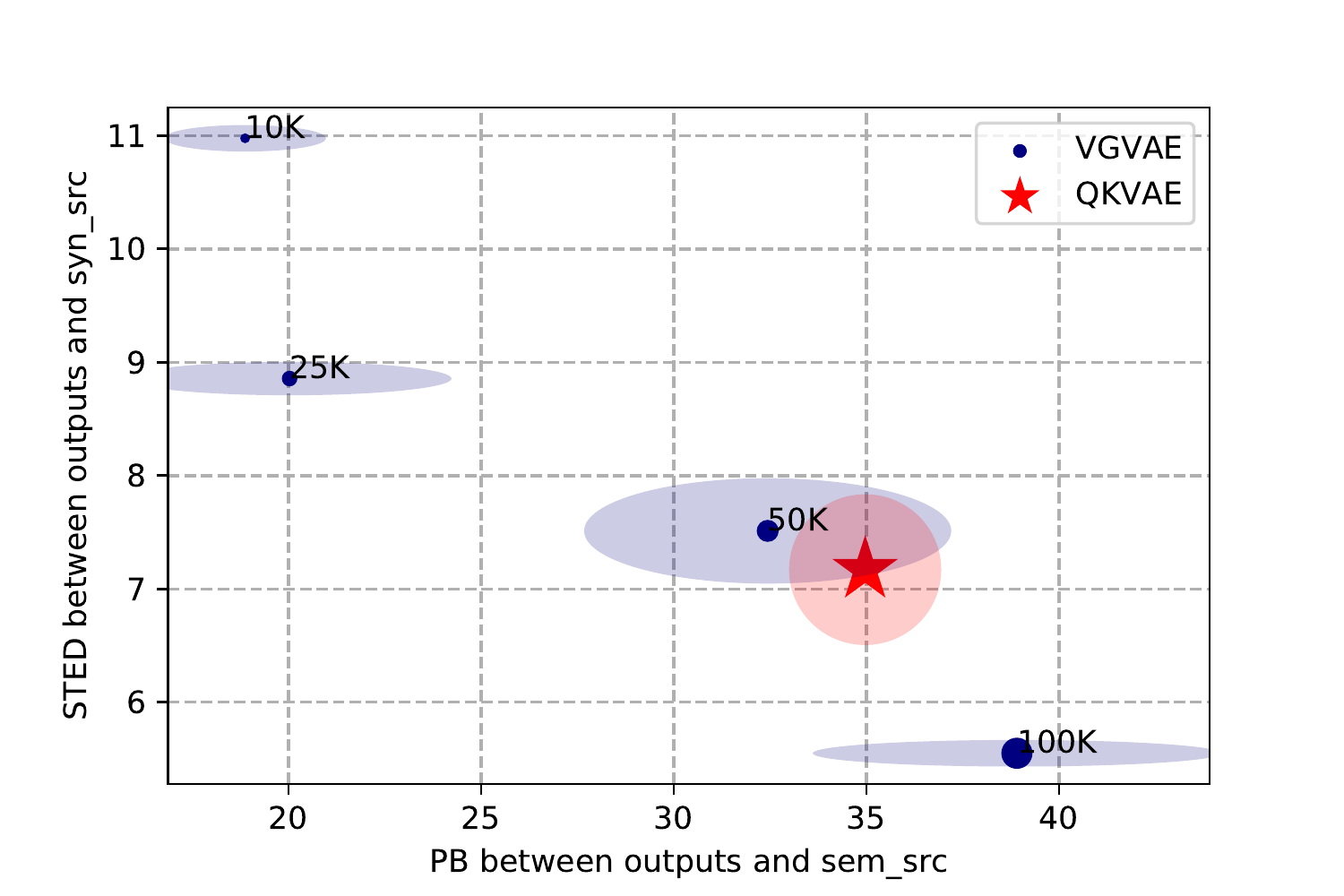}
  \end{adjustbox}
  \end{minipage}
  \caption{\gfcmt{I regenerated this figure with averages over 5 runs and standard deviations. }Plotting STED w.r.t \emph{syn\_ref} and the PB cosine similarity w.r.t \emph{sem\_ref} for VGVAE with different amounts of labeled data and for QKVAE.
   Points are scaled proportionally to the amount of training data. \gfcorr{$\emptyset$}{The vertical and horizontal diameters of each ellipse are equal
   to the standard deviation of the associated data points and axes.}}
  \label{fig:STEDvsPB}
\end{figure} 
In Figure~\ref{fig:STEDvsPB}, we plot for QKVAE and instances of VGVAE the \emph{STED} of their outputs w.r.t
 \emph{syn\_src} and the \emph{PB} of these outputs w.r.t \emph{sem\_src}. \gfcorr{$\emptyset$}{All values are averages over 5 runs, with standard
 deviations plotted as ellipses.}
 Figure~\ref{fig:STEDvsPB} shows that to outperform QKVAE on  \gfcorr{semantic transfer, VGVAE needs more than 25K labeled samples. 
To outperform our model on both transfer metrics, it needs more than 50K samples.}{syntactic and semantic transfer,  VGVAE needs more than 50K labeled samples.} 

\section{Discussion and conclusion}
In Table \ref{tab:qualiRes}, we display example outputs of SynPG, VGVAE, and QKVAE along with their syntactic sources, 
semantic sources, and targets. We generally observed that the outputs of QKVAE range from paraphrases (line \gfcorr{1}{6}) to broadly 
related sentences (line \gfcorr{2}{3}). 
\gfcorr{It also displays some flexibility in terms of lexical choices which allows it to 
introduce relevant words which were not in the semantic source (\textit{e.g.} \emph{killed} in line 3).
 We also observed this quality for VGVAE, and we attribute it to the smoothness of the latent space of VAEs which places 
 coherent alternative lexical choices in the same vicinity. Moreover, }{
   As was shown by our quantitative results, outputs from VAE-based models (VGVAE and QKVAE) share relatively few lexical items
   with the semantic input. This can be seen in the qualitative examples where they often swap words in the semantic
   source with closely related words (\textit{e.g.} "armored division" to "military force" in line 1, or "lunch boxes" to 
   "snacks" in line 2). We attribute this quality to the smoothness of the latent space of VAEs which places 
   coherent alternative lexical choices in the same vicinity. The examples above also show that
 } our model is capable of capturing and transferring various syntactic characteristics such
  as the passive form (line \gfcorr{3}{1}), the presence of subject-verb inversion (line\gfcorr{ 1}{s 3, 4, and 7}),
   or interjections (\gfcorr{also line 1}{lines 4 and 6}).



We presented QKVAE, an unsupervised model which disentangles syntax from semantics without syntactic or semantic information.
Our experiments show that its latent variables effectively position sentences in the latent space according to these attributes.
Additionally, we show that QKVAE displays clear signs of disentanglement in transfer experiments.
Although the semantic transfer is moderate, syntactic transfer with QKVAE is competitive with SynPG, one of its supervised counterparts.
Finally, we show that VGVAE, a supervised model, needs more than 50K samples to outperform QKVAE on both syntactic and semantic transfer.

   We plan to extend this work in \gfcorr{two}{three} directions: 
\text{i)} Finding ways to bias representations of each $z^{sem}_l$ towards understandable concepts; \textit{ii)} Applying QKVAE 
to non-textual data since it is data agnostic (\text{e.g.} to rearrange elements of a visual landscape.)\gfcorr{$\emptyset$}{; \textit{iii)} Investigating the behavior of QKVAE on other languages.}  



\section*{Acknowledgments}
This work is supported by the PARSITI project grant (ANR-16-CE33-0021) given by the French National Research Agency (ANR), 
the \emph{Laboratoire d’excellence “Empirical Foundations of Linguistics”} (ANR-10-LABX-0083), as well as the ONTORULE project.
 It was also granted access to the HPC resources of IDRIS under the allocation 20XX-AD011012112 made by GENCI.
\bibliography{references, bib2}
\bibliographystyle{acl_natbib}

\appendix

\clearpage

\begin{table}[b!]\small
  \centering
  \begin{tabular}{l c c}
  \hline
  &$z^{sem}\uparrow$ & $z^{syn}\downarrow$\\
  \hline\multicolumn{3}{c}{\textit{Supervised Models}}\\\hline
  VGVAE & 99.0& 16.4\\
  SynPG & 91.6& 31.2\\
  \hline\multicolumn{3}{c}{\textit{Unsupervised Models}}\\\hline
  Optimus & 89.4 & -\\
  ADVAE & 41.0& 40.3\\
  QKVAE & 86.7& 27.0\\
  \hline
  \end{tabular}
  \caption{\label{enc_res_dev} \gfcmt{I replaced the results of ADVAE and QKVAE with the 5 runs averages here. }
  The probability*100 that an embedding places a target sentence closer to its semantic source than it is to its syntactic source 
  in the embedding space. (development set results)}
  \end{table}
\begin{table}[b!]\small
\centering
\begin{tabular}{p{0.045\textwidth} c| c|| c| c|| c| c}
\hline
&\multicolumn{2}{c}{\textit{sem\_src}} & \multicolumn{2}{c}{\textit{syn\_src}}& \multicolumn{2}{c}{\textit{target}}\\
& \emph{M}$\uparrow$&\emph{PB}$\uparrow$
& \emph{M}$\downarrow$&\emph{PB}$\downarrow$
& \emph{M}$\uparrow$& \emph{PB}$\uparrow$ \\
\hline
\multicolumn{7}{c}{\textit{Control and Reference baselines}}
\\
\hline
\textit{sem\_src}    & 100 & 1.0 &  7.4& 0.13& 27.4&  0.82\\
\textit{syn\_src}    & 7.4& 0.13&  100 & 1.0 &  12.0& 0.16\\
Optimus & 13.00& 0.35& 13.4& 0.34$^\dag$&  10.5 & 0.32\\
\hline
\multicolumn{7}{c}{\textit{Supervised Models}}
\\
\hline
VGVAE    & 18.3& 0.58&  15.2& 0.17& 23.0 & 0.57\\
SynPG  & 47.6 & 0.86& 7.8& 0.11& 24.4& 0.73\\
\hline
\multicolumn{7}{c}{\textit{Unsupervised Models}}
\\
\hline
ADVAE    & 9.0& 0.20& 8.1& 0.17& 7.7& 0.19\\
QKVAE    & 13.4& 0.36&  11.3& 0.19 & 12.9 &0.35\\
\hline

\end{tabular}
\caption{\label{sem_res_dev} \gfcmt{I replaced the results of ADVAE and QKVAE with the 5 runs averages here. }
Semantic transfer results (development set results)
}
\end{table}

\begin{table*}[b!]\small
\centering
\begin{tabular}{l c c c|| c c c|| c c c }
\hline
&\multicolumn{3}{c}{\textit{sem\_src}} & \multicolumn{3}{c}{\textit{syn\_src}}& \multicolumn{3}{c}{\textit{target}}\\
&\emph{STED}$\uparrow$ & \emph{TMA2}$\downarrow$ & \emph{TMA3}$\downarrow$ & 
\emph{STED}$\downarrow$ & \emph{TMA2}$\uparrow$ & \emph{TMA3}$\uparrow$ &
\emph{STED}$\downarrow$ & \emph{TMA2}$\uparrow$ & \emph{TMA3}$\uparrow$ \\
\hline
\multicolumn{10}{c}{\textit{Control/Ceiling baselines}}
\\ 
\hline
\textit{sem\_src} &0.0 & 100 & 100 & 11.9& 46.4& 6.8 & 10.9& 47.0&7.3 \\
\textit{syn\_src} & 11.9& 46.4& 6.8& 0.0& 100& 100& 6.0& 81.6& 45.0\\
Optimus & 9.7& 58.2& 20.6& 9.2$^\dag$& 61.6$^\dag$& 22.6$^\dag$& 9.9& 59.6& 18.4\\
\hline
\multicolumn{10}{c}{\textit{Supervised Models}}
\\
\hline
VGVAE    & 11.9& 45.4& 6.8& 3.2& 84.2& 58.2& 6.7& 77.6& 39.0\\
SynPG & 9.3& 49.4& 21.4& 12.2& 73.0& 12.2& 12.2& 68.6&13.0\\
\hline
\multicolumn{10}{c}{\textit{Unsupervised Models}}
\\
\hline
ADVAE    & 10.1& 53.4& 18.6& 9.8$^\dag$& 55.0$^\dag$& 17.4$^\dag$& 10.5& 52.8& 15.4\\
QKVAE    & 11.4& 45.0& 9.1& 6.8& 66.4& 37.4& 8.6& 63.0&26.9\\
\hline

\end{tabular}
\caption{\label{syn_res_dev}
\gfcmt{I replaced the results of ADVAE and QKVAE with the 5 runs averages here. }Syntactic transfer results (development set results)
}
\end{table*}
\section{Results on the development set}
\label{DEVRES}
\begin{figure}[!h]
  \hskip -0.6cm
  \begin{minipage}[b]{\textwidth}
    \begin{adjustbox}{minipage=\textwidth,scale=0.59}
    \hspace{ 1cm} \includegraphics[trim={0.7cm 0.2cm 1.5cm 1.2cm},clip] {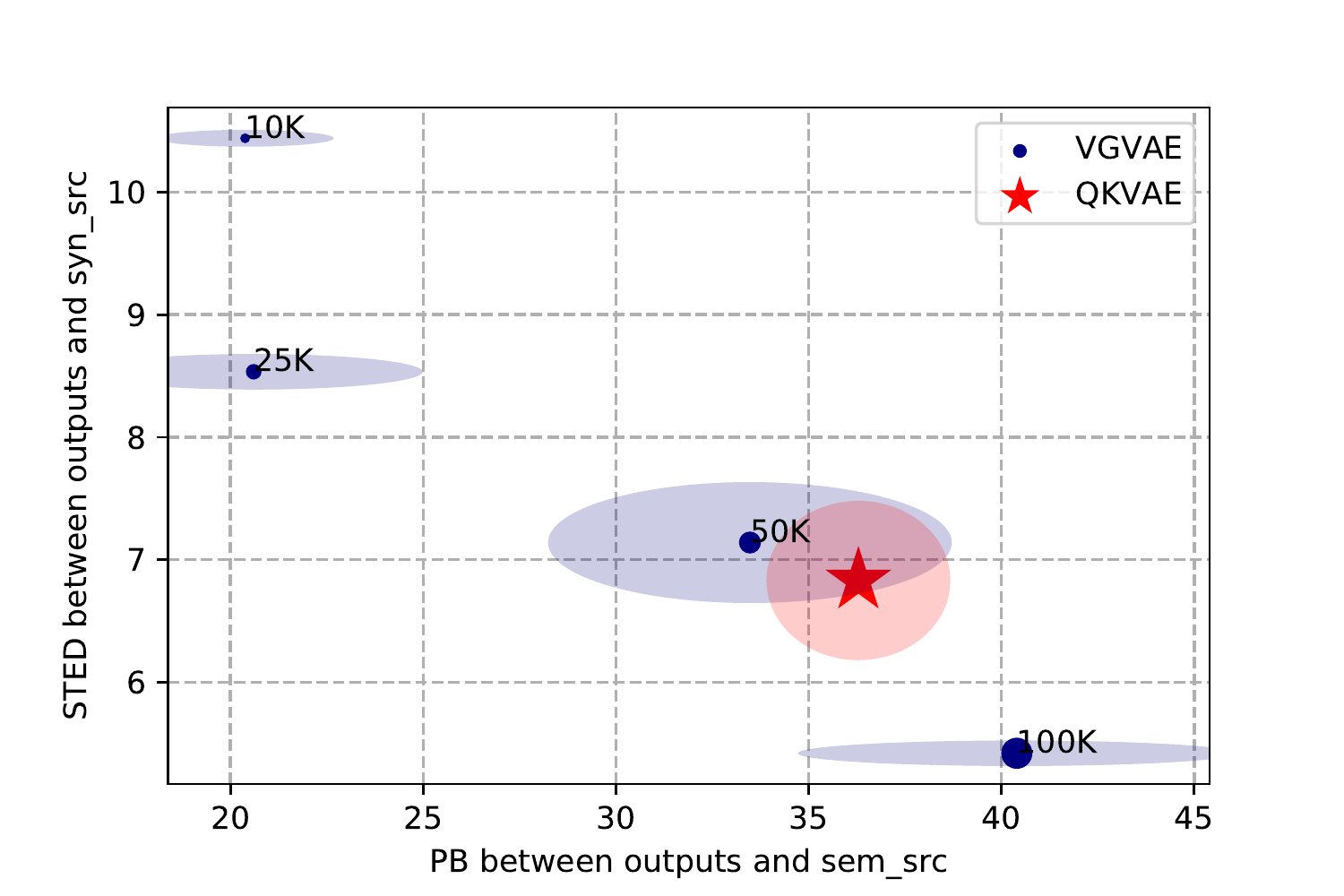}
  \end{adjustbox}
  \end{minipage}
  \caption{\gfcmt{I added this figure since these results were not included for the dev set. }Plotting STED w.r.t \emph{syn\_ref} and the PB cosine similarity w.r.t \emph{sem\_ref} for VGVAE with different amounts of labeled data and for QKVAE.
   Points are scaled proportionally to the amount of training data. \gfcorr{$\emptyset$}{The vertical and horizontal diameters of each ellipse are equal
   to the standard deviation of the associated data points and axes.}
   }
  \label{fig:STEDvsPBdev}
\end{figure}

We hereby display the scores on the development set.
The encoder scores concerning the specialization of latent variables are in Table \ref{enc_res_dev}, while the transfer scores
 are in Table \ref{sem_res_dev} for semantics, and Table \ref{syn_res_dev} for syntax. The values on the development set concerning
 the comparison of QKVAE with VGVAE trained on various amounts of data is in Figure \ref{fig:STEDvsPBdev}.

\section{Hyper-parameters}
\label{HPAPPEN}
\paragraph{Hyper-parameter values}
The $\beta$ weight on the $\KL$ divergence is set to 0.6 \gfcorr{$\emptyset$}{for $z^c$ and to 0.3 for $z^s$}, and 
the $\lambda$ threshold for the Free-Bits strategy is set to \gfcorr{0.1}{0.05}.
$\KL$ annealing is performed between steps 3K and 6K for $z^{sem}$, and between steps 7K and 20K for $z^{syn}$.
The model is trained using Adafactor \cite{Shazeer2018Adafactor:Cost}, a memory-efficient version of Adam \cite{Kingma2015}.
Using a batch size of 64, we train 
for \gfcorr{70}{40} epochs, which takes about \gfcorr{4 days}{30 hours} on a single Nvidia GEForce RTX 2080 GPU. \gfcorr{$\emptyset$}{We use 4 layers for both Transformer encoders and decoders.
The encoders (resp. decoders) are initialized with parameters from the 4 first layers (resp. 4 last layers) of BART encoders (resp. decoders).}
 In total, our model uses \gfcorr{291}{236}M parameters. 

 \paragraph{Manual Hyper-parameter search} \gfcorr{$\emptyset$}{Given that the architecture for Transformer layers  is fixed by BART,
  we mainly explored 3 parameters: number of latent variables $L$, number of Transformer layers, values for $\beta$. Our first 
  experiments have shown that setting $L$ to 8 or 16 does not yield good results, which is probably due to the fact that a high 
  $L$ raises the search space for possible arrangements of values with keys, and consequently makes convergence harder. Concerning
  the number of layers, we observed that results with the full BART model (6 layers) have high variance over different runs. Reducing
  the number of layers to 4 solved this issue. In regards to $\beta$, we observed that it must be $0.6$ or less for the model to 
  produce adequate reconstructions and that it is beneficial to set it slightly lower for $z^{syn}$ than for $z^{sem}$ so
   as to absorb more syntactic information with $z^{syn}$. }

\end{document}